# Figure Descriptive Text Extraction using Ontological Representation

Gilchan Park[1], Julia Rayz[2], Line Pouchard[1]

[1]Brookhaven National Laboratory, [2]Purdue University
{gpark, pouchard}@bnl.gov, jtaylor1@purdue.edu

## Abstract

Experimental research publications provide figure form resources including graphs, charts, and any type of images to effectively support and convey methods and results. To describe figures, authors add captions, which are often incomplete, and more descriptions reside in body text. This work presents a method to extract figure descriptive text from the body of scientific articles. We adopted ontological semantics to aid concept recognition of figure-related information, which generates human- and machine-readable knowledge representations from sentences. Our results show that conceptual models bring an improvement in figure descriptive sentence classification over word-based approaches.

## Introduction

Scientific literature can be viewed as a plethora of documents written in natural language where the agglomeration of well-cogitated knowledge is inextricably intertwined. The sheer volume of knowledge source has the great potential to derive valuable information when text-mined and untangled. Scientific literature mining studies have been taking place across disciplines to automate knowledge acquisition process, or solve problems dealing with a large amount of textual data (Carvaillo et al. 2019; Kim et al. 2017b; Kveler et al. 2018; Tchoua et al. 2016). One useful application is automatic relevant information retrieval. Given the fast-growing digitized research papers, the establishment of data collection from literature facilitates scientific discovery and reduces manual labor and human error.

One of the important contents in scientific articles is figures. According to Futrelle (2004), approximately 50% of the contents in biology articles are figure relevant, and the most informative contents of chemistry papers to chemists were the authors and the figures. Schematic representations are effective to conceptualize the notion of methods and deliver experimental results in a straightforward and comprehensible way. For this reason, readers tend to first identify prominent figures to quickly comprehend articles so that they can find articles of interest among a large amount of complicated information (Pain 2016). In particular, when experimental outputs are represented in a universal format in a domain, figures are the most effective means to convey ideas. Additionally, such standardized figures can be used to group articles by methods and results, which is useful for scientists to find research patterns and trends. To explain figures, authors provide captions as explanatory notes. Since figure captions are mostly concise and simple, information they contain is often insufficient, and the fully descriptive information resides in body texts. Therefore, to fully understand figures, not only the captions but also relevant body text should be digested (Yu et al. 2009).

The proposed approach aims to extract figure descriptions represented in text. Specifically, the method targets content that surrounds figure referring sentences (e.g., 'Fig.1 represents'). We assume that figure referring sentences are informative but are not complete, and further information can be found in their surrounding text. In this paper, we define figure descriptive text as a set of sentences that directly describe or explain figures. The hypothesis is that text describing objects will have distinguishable concepts from other explanatory text. For instance, graphical symbols, the speed of a change, movement and shape of entities can be common concepts of descriptive text. To perform the experiments, we designed a conceptual model to identify figure descriptive sentences by meaning-based representations.

## Related Work

There have been some efforts to identify figure relevant texts in scientific publications.

Agarwal and Yu (2009) developed a figure summarization system, *FigSum*, which extracted contents pertinent to figures from journal articles in the biomedical domain. Their algorithm calculates sentence scores by word similarity with figure caption and representative terms to article theme, and by which it selects the most figure relevant sentences.

Ramesh, Sethi, and Yu (2015) further improved *FigSum* by incorporating figure reference features, which take into account whether a sentence is a figure referring sentence or belongs to the same paragraph of figure referring sentences. They evaluated the system on 19 full-text biomedical articles and reported that figure reference features played the most important role to identify figure relevant texts.

Takeshima and Watanabe (2012) suggested a weight propagation method to measure not only positional significance of sentences (i.e., distance from figure referring sentence), but also word weights in sentences for sentence similarity. The weights of words and sentences are calculated and updated by each other, which intends to assign high values to similar sentences about figure explanation. The system evaluation showed around 76% precision for figure relevant text extraction.

The previous works utilized structural properties and superficial patterns in text for sentence similarity. On the other hand, little effort was made to measure semantic relations between sentences. To advance semantic analysis and generate more meaningful and human-comprehensible outcomes, we introduce a concept-based model based on ontological representation.

An ontology is a set of concepts and properties that define attributes and semantic relations of concepts to describe a specific domain or world (Nirenburg and Raskin 2004). An ontological system is built upon rules of semantic components and common sense by human efforts to reflect human reasoning and world knowledge. With the recent advances in statistical analysis techniques, in particular neural network models in natural language processing (Young 2018), text mining research has been vigorously employing such methods to process high-throughput data and improve textual pattern recognition. In many cases, contextual information to identify semantic relations that machines use in such schemes is limited to the given data. The ontological system supplies semantic knowledge that goes beyond the scope of texts to the machines by incorporating human understanding.

# Experiments

This section describes figure descriptive sentence extraction process for a domain of X-ray absorption spectroscopy.

## Target Domain & Dataset

The target domain for the experiments is X-ray absorption spectroscopy (XAS). XAS is a well-known technique to characterize structural properties of elements using synchrotron radiation, and a myriad of papers has reported XAS results for elements in different materials. XAS results are represented as graphs in articles, and they convey valuable information for experimenters. To collate relevant articles, we utilized Elsevier's TDM (Text and Data Mining) services. Elsevier provides RESTful (Representational state transfer) APIs to search for articles of interest and scrape them. Basically, the search mechanism is based on keyword matching. We chose four keywords with XAS domain experts – XAFS (X-ray Absorption Fine Structure), EXAFS (Extended X-ray Absorption Fine Structure), XANES (X-ray Absorption Near Edge Structure), NEXAFS (Near Edge X-ray Absorption Fine Structure). XAFS (= EXAFS) and XANES (= NEXAFS) are the most common XAS techniques, and most XAS experimental articles contain either of those terms. Using these selected keywords, the article scraper of our system has collected 40,561 papers from the Elsevier archive.

The raw articles are stored in XML (eXtensible Markup Language) format. The text pre-processor segmented XML data into elements by publisher-specific XML schema. We applied a domain-specific parsing tool, ChemDataExtractor (Swain and Cole 2016), to parse texts and identify chemical named entities. ChemDataExtractor is built upon natural language processing pipeline and machine learning algorithms, and the toolkit produced outstanding results in recognition of chemical structures and properties (Court and Cole 2018; Kim et al. 2017). Body texts are parsed in a sentence boundary. The parsed data were reformatted into a unified JSON format. The elements of JSON files are *UID (unique identifier), publisher, article type, title, year, author, keywords, abstract, body text, figures*.

In the earlier work (Park and Pouchard 2019), we built a simple heuristic-based model for taxonomic classification of papers by 30 transition metals and types of XAS edges (K/L/M-edge) to provide scientists with filtered list of papers. The model was designed to analyze figure captions for the classification. However, it revealed that captions do not always have complete information, and the relevant body texts need to be further examined to increase classification accuracy.

## Sentence selection

Capturing textual descriptions of figures in body texts starts with setting the range of textual segments to be considered. We differentiate between two types of sentences: figure referring sentences and their neighboring sentences.

We define figure referring sentences as sentences having the term – *fig(s), figure(s)* followed by a number (e.g., Fig.3, Figure S1, Figs 1-2). Figure referring sentences usually contain associated information to figures including figure descriptions, but the information is often insufficient, and presumably additional text surrounding them can fill the lack of information. Therefore, sentences that surround figure referring sentences (hereafter called "neighboring sentence") were chosen as the candidates to be examined. Among the neighboring sentences, we are interested in finding sentences directly describing figures, whether the word figure is present or not (e.g., *"All the spectra show a prominent resonant microwave absorption signal (Hr) around 3000 G.", "In this calculation, we set the thickness of the oxide layer as 15 nm and the fraction of $Cu_2O$ (CuO) as 64% (36%), for the case of graphene-coated copper."*)

To define a range of neighboring sentences, sentence window was set to two, which means preceding and succeeding two sentences of a figure referring sentence were the text segment to be examined. If a considered neighboring sentence is another figure referring sentence, then the sentence is discarded as a neighboring sentence. In addition, the sentence window is only effective in the same paragraph where the referring sentence exists. In general, a paragraph conveys a single idea and consists of coherent sentences to the idea. Thus, it is reasonable to assume that sentences from different paragraphs are irrelevant to a description of a figure. If a figure referring sentence is at the beginning or end of a paragraph, its preceding or succeeding sentences are excluded from the neighboring sentences respectively.

Figure referring sentences have a unique identifier to be readily detected; whereas, neighboring ones describing figures do not have noticeable clues. The main task is to discover distinct features, in particular conceptually, to determine whether a sentence contains a figure description.

**Knowledge Representation**

To create an ontology for figure descriptive text detection, we adopted Ontological Semantic Technology (OST) (Hempelmann, Taylor, and Raskin 2010). OST conducts semantic processing for natural language text based on ontological semantics. OST consists of ontology and lexicon. The ontology is a language independent component where concepts under EVENT and OBJECT are interconnected by a set of properties such as subsumption relations (IS-A: hierarchical connectivity) and mereological relations (PART-WHOLE) (Raskin et al. 2010). Ontology has a graph structure in which nodes and edges indicate concepts and properties respectively. Lexicon is a machine-processable dictionary for a specific natural language that contains word senses mapping to corresponding concepts in the ontology. OST takes a sentence as input and finds the best combination of word senses of sentence components taking into consideration their relations permitted by the ontology. As a result, OST generates a resulting graph of the most appropriate concept set, called TMR (Text Meaning Representation), which is both machine- and human-readable knowledge representation for an input sentence.

To create a TMR for a sentence, this work takes into account three main components in a sentence – *subject*, *verb* and *object*, and their modifiers such as *adjective* and *adverb*. These components were recognized by the dependency parser in spaCy natural language processing tool. For domain-specific terms related to chemical objects, the chemical named entity recognizer in ChemDataExtractor eased the difficulty in handling numerous chemical names (e.g., gold, sulfate, $TiO_2$, FeS – the concept CHEMICAL).

The current model solely utilizes information within a sentence, and thereby missing or omitted information often exist. This may hinder finding a proper path, in particular when a subject or an object is a property of an unrevealed concept. To alleviate this problem, UNKNOWN entities are temporarily added to the ontology graph to complete a path (see Figure 2 as example). The example shows that dependency parser finds syntactic relations between words; these relations are then passed to semantic parser that takes into account lexemes and their ontological representation; ontological representation is used to produce TMRs.

**Concept selection.** To find representative concepts in the domain, we first found frequent words in the dataset. We analyzed figure referring sentences from 9,164 articles and found most common verbs including *show, exhibit, display, depict, illustrate, demonstrate, visualize, describe, reveal, plot, give, compare, observe, obtain, increase, decrease*. We scrutinized the verbs' dependents (subjects, object, modifiers) to create figure descriptive concepts. As a modifier, adjective and adverbs are frequently used to describe figures including charts, graphs, and diagrams. SHAPE, COLOR, SAMENESS, SIZE, and QUANTITY are some of the most common descriptive properties. Table 1 lists some of the properties for adjectives and adverbs. Value column indicates the literal values of properties. For example, *straight* is the value of the property SHAPE in the phrase *straight line*.

Table 1. Examples of properties for adjectives and adverbs.

| **Property** | **Value** | **Lexemes** |
|---|---|---|
| SHAPE | polygonal, straight, round, … | triangular, rectangular, linear |
| SPEED | fast, average, slow | rapid, slow |
| DIRECTIONALITY | horizontal, vertical | flat, vertical |
| DESIRABILITY | high, medium, low | good, bad |
| SAMENESS | high, medium, low | same, different |
| QUANTITY | small, neutral, large | few, little, many, numerous, some |
| ORDER | preceding, next, last | prior, subsequent, last |
| SAMENESS | high, medium, low | identical, dissimilar, different |

**Expanding the scope of verbs.** To include more verbs than the selected verbs, we leveraged WordNet (Miller 1995). WordNet is a computational resource of a combination of dictionary and thesaurus built by human efforts and basically groups words into synonym sets (hereafter called synsets). Specifically, when a verb is not defined in the OST lexicon, its synonym is used to identify the verb's concept based on relations with its dependents (subject and object). However, word senses defined in WordNet are often too fine-grained, and processing all the synsets of senses causes a computational overhead. To filter out irrelevant synonyms, we incorporated a word embedding model where context similarity of words somewhat disambiguates word senses although the disambiguation is imperfect and limited to the given data. The rationale behind combining the two approaches is that disambiguation process for the synonyms defined by humans is assisted by word embeddings' contextual similarity that helps reduce the number of candidate senses, which will be eventually verified by the OST. The adopted word embedding technique is Word2Vec (Mikolov et al. 2013), and a model was trained on 369,287 figure referring sentences. When encountering an undefined verb, the OST checks common synonyms between the verb's synsets and the top 20 most similar words of the verb by the word embeddings to find the concept of the verb.

Figure 1 describes the overall process of TMR generation. In the example, the word *position* is not a concept, but a property of object. However, the object is not known in the sentence. To complete a path, two UNKNOWN entities were added between show and position – one was the concept of the unknown object and the other was the unknown value of GEOMETRIC-ASPECT property.

### Score function

Scores of TMR graphs are measured by calculation of weights of elements in graphs. In a TMR graph, end nodes (concepts) and edges (properties) are regarded as the most significant since they are direct representations of words and semantic relations, and the significance of elements diminishes as the distance from words increases. To apply this notion, weights of concepts and properties (shown in equation 1) are reduced to distance from the end node to power of 2, and the weight of element is normalized by total weights of each category (i.e., concept or property). During the calculation of concept weights, top level elements, EVENT and OBJECT, are ruled out from the concept list. This is because they appear in all TMRs due to their top ranked position in the concept hierarchy, which allocates high weights to them although they do not imply as much semantic information as other concepts in a TMR. With regard to properties, the CASE-ROLE and CAUSALITY properties, AGENT, THEME, THEME-INFORMATION, INSTRUMENT, CAUSED-BY, are basic relations between verbs and their dependents, and they appear in all TMRs. Due to their omnipresence, CASE-ROLE and CAUSALITY properties do not provide distinguishable features for figure descriptions unlike ATTRIBUTE properties for modifiers in Table 1. Thus, they were excluded from the property list to put more weights on ATTRIBUTE properties. We also did not consider IS-A properties which are nothing

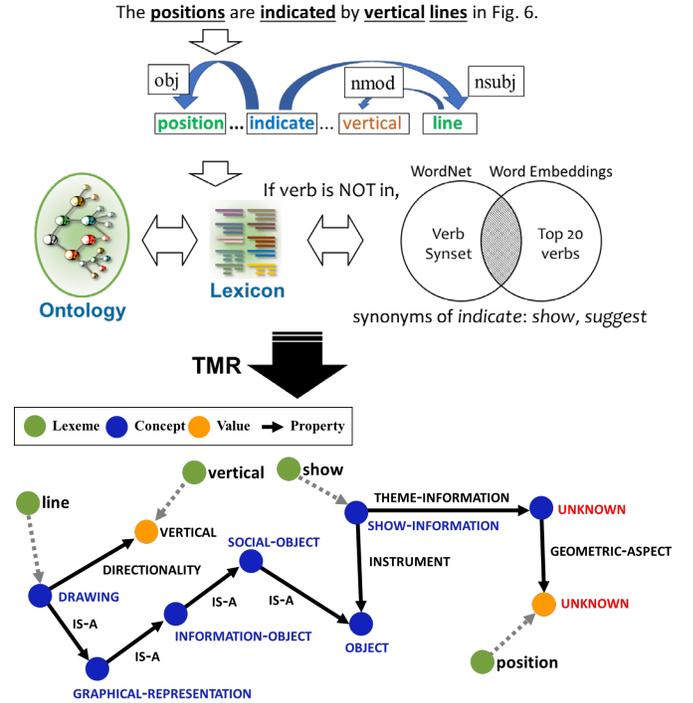

Figure 1. The process of TMR generation.

more than hierarchical relations of concepts (ancestors and successors).

To set initial weights of concepts and properties and generates a threshold value for classification, we used TMRs generated by figure referring sentences. The total number of the TMRs was 9,087, and 270 concept and 39 properties were found. Elements are calculated by equation 1, and $\alpha$ is a normalization factor.

$$Weight(e) = \alpha \left\{ \sum_{e \in E_f} \frac{1}{distance(e)^2} \right\} \quad (1)$$

$$\alpha = \frac{1}{\sum_i^N Weight(i)} \quad (2)$$

where $E_f$ is the set of all elements of either concepts or properties in figure referring sentences; *distance* indicates the distance (path) from a lexeme to a known concept (e.g., 1, 2, 3, 4…). The weight of element is normalized by the total sum of weights of its category (concept or property).

For each sentence, concepts and properties are summed by the normalized scores considering their distance value (shown in equation 3). To evaluate the model, the sentence classification threshold value is set by equation 4.

$$Weight(s) = \sum_{e \in E_s} \frac{Weight(e)}{distance(e)^2} \quad (3)$$

$$Threshold = \frac{1}{|S_f|} \lambda \sum_{s \in S_f} Weight(s) \quad (4)$$

where $E_s$ is the set of all elements in a sentence. $S_f$ is the set of all figure referring sentences, and the threshold is set to the mean of the weights of all figure referring sentences, which is further adjusted by a hyperparameter $\lambda$. Figure 2 shows the distance values of concepts and properties in TMR illustrated in Figure 1, where pink colored circles denote distance of property and yellow colored circles denotes distance of concepts. Table 2 illustrates the calculation of weights for the TMR graph.

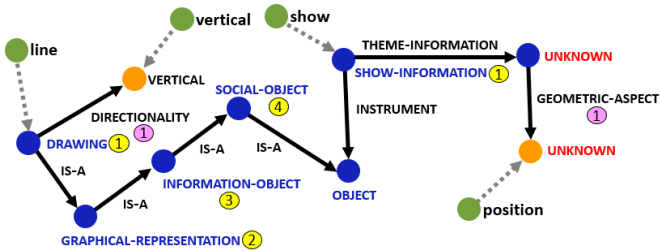

Figure 2. The distance of concepts and properties.

Table 2. The process of TMR weight calculation.

| Element (distance) | Initial weight/distance$^2$ |
| --- | --- |
| DRAWING (1) | $0.0298/1^2 = 0.0298$ |
| GRAPHICAL-REPRESENTATION (2) | $0.2332/2^2 = 0.0583$ |
| INFORMATION-OBJECT (3) | $0.1086/3^2 = 0.0121$ |
| SOCIAL-OBJECT (4) | $0.022/4^2 = 0.0014$ |
| SHOW-INFORMATION (1) | $0.2052/1^2 = 0.2052$ |
| DIRECTIONALITY (1) | $0.0055/1^2 = 0.0055$ |
| GEOMETRIC-ASPECT (1) | $0.1154/1^2 = 0.1154$ |
| Sentence weight | 0.4277 |

**Model Evaluation**

We assessed the model performance in a dataset consisting of 100 figure descriptive sentences and 100 non-figure descriptive ones. The sentences were randomly chosen from neighboring sentences of figure referring sentences in 137 XML articles, containing 22,352 sentences. The model generated TMRs of these sentences, which were classified into two classes – figure description and non-figure description by the equation 4. If a TMR score is greater than the threshold, the sentence is considered a figure descriptive sentence. To set the hyperparameter $\lambda$ value, we compared different $\lambda$ values to identify figure referring sentences (not figure descriptive ones that were used to test the methodology). Table 3 shows the detection accuracy on all figure referring sentences with different $\lambda$ values.

Table 3. Accuracy on figure referring sentences with different $\lambda$

| Threshold ($\lambda$) | Accuracy |
| --- | --- |
| 0.0407 (0.1) | 0.9896 |
| 0.1221 (0.3) | 0.862 |
| 0.2035 (0.5) | 0.7994 |
| 0.2848 (0.7) | 0.6892 |
| 0.3662 (0.9) | 0.5578 |
| 0.6104 (1.5) | 0.1692 |

It can be seen that the classification accuracy increases as $\lambda$ decreases, but a low cutoff threshold also escalates false positives in classifying neighboring sentences. As a trade-off point, we selected 0.5 for the $\lambda$ value which showed a reasonable accuracy of 80%. The classification test on the labeled 200 neighboring sentences resulted in 0.82 $F$-score with the $\lambda$ value of 0.5. We tested other values to compare the performance and found that the model indeed performed the best with $\lambda$ set at 0.5 (shown in Table 4).

Table 4. Performance on neighboring sentences with different $\lambda$

| Threshold ($\lambda$) | Accuracy | $F$-score |
| --- | --- | --- |
| 0.0407 (0.1) | 0.52 | 0.3839 |
| 0.1221 (0.3) | 0.735 | 0.7314 |
| **0.2035 (0.5)** | **0.82** | **0.8199** |
| 0.2848 (0.7) | 0.74 | 0.7348 |
| 0.3662 (0.9) | 0.645 | 0.6059 |
| 0.6104 (1.5) | 0.535 | 0.4067 |

We further tested our model on articles in PDF (Portable Document Format) format which has become the dominant file type for research publications. Unlike XML articles PDF articles do not have tags or delimiters to further split text into sections, paragraphs, and special segments such as tables, math formulas, figure captions. Therefore, when segmented text data is not available, for instance figure captions for XAS classification, extracting textual information of interest from article is more important. We used the PyMuPDF parsing tool to preprocess PDF files. PDF articles from Institute of Physics were used for testing. As in the test on XML articles, we selected 100 figure descriptive sentences and 100 non-figure descriptive ones from neighboring sentences of figure referring sentences in 157 articles, containing 36,542 sentences. The model on the PDF dataset produced 0.78 $F$-score with the $\lambda$ value of 0.5, which also performed the best among the other values.

To compare our conceptual model with a bag-of-words approach, we tested three well-known machine learning algorithms – logistic regression (LR), support vector machine (SVM), and random forest (RF). The three algorithms were trained on 400 manually classified sentences (200 XML and 200 PDF) used for the experiments of the concept model. Figure 3 presents the results of 10-fold cross validation (CV) of the three algorithms along with the performance of conceptual models for comparison. The conceptual models outperformed bag-of-words models, and the performance difference was statistically significant ($\alpha = .05$).

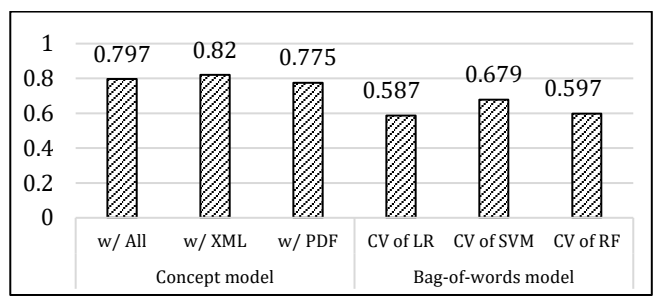

Figure 3. *F*-scores of Concept and Bag-of-words models.

## Discussion and Future Work

This research attempted to find figure descriptive text from scientific literature with an ontological model designed to conceptualize words and find figure descriptive concepts. Human-engineered semantic analysis of text successfully captured distinguishable concepts for figure descriptive sentences and yielded a higher sentence classification accuracy than superficial feature-based approaches.

For future work, we will expand the scope of sentence components and handle context information between sentences to improve conceptualization and classification performance. This work is an initial step of meaning-based figure relevant text extraction and focused on descriptive text of objects. We will further capture more informative textual information beyond being descriptive. To process this in a systematic way, it is required to set the level of relatedness between textual information and figures and define the rules with domain experts. We will explore ways to gather and structuralize experts' knowledge and to compare semantic contents similarity between sentences. The result of this collaborative work will eventually assist the domain scientists to efficiently gain supplementary information while preparing experiments or analyzing results.

## Acknowledgements

This work has been authored by employees of Brookhaven Science Associates, LLC operated under Contract No. DESC0012704. The authors gratefully acknowledge the funding support from the Brookhaven National Laboratory under the Laboratory Directed Research and Development 18-05 FY 18-20.